\documentclass{article}

\usepackage{microtype}
\usepackage{graphicx}
\usepackage{subcaption}
\usepackage{booktabs} % for professional tables

\usepackage{hyperref}

\usepackage[preprint]{icml2026}

\usepackage{amsmath}
\usepackage{amssymb}
\usepackage{mathtools}
\usepackage{amsthm}

\usepackage[capitalize,noabbrev]{cleveref}

\theoremstyle{plain}
\newtheorem{theorem}{Theorem}[section]
\newtheorem{proposition}[theorem]{Proposition}
\newtheorem{lemma}[theorem]{Lemma}
\newtheorem{corollary}[theorem]{Corollary}
\theoremstyle{definition}
\newtheorem{definition}[theorem]{Definition}
\newtheorem{assumption}[theorem]{Assumption}
\theoremstyle{remark}
\newtheorem{remark}[theorem]{Remark}

\usepackage[textsize=tiny]{todonotes}

\icmltitlerunning{Bag of Coins}

\begin{document}

\twocolumn[
\icmltitle{Bag of Coins: A Statistical Probe into Neural Confidence Structures}

  % It is OKAY to include author information, even for blind submissions: the
  % style file will automatically remove it for you unless you've provided
  % the [accepted] option to the icml2026 package.

  % List of affiliations: The first argument should be a (short) identifier you
  % will use later to specify author affiliations Academic affiliations
  % should list Department, University, City, Region, Country Industry
  % affiliations should list Company, City, Region, Country

  % You can specify symbols, otherwise they are numbered in order. Ideally, you
  % should not use this facility. Affiliations will be numbered in order of
  % appearance and this is the preferred way.
  \icmlsetsymbol{equal}{*}

 \begin{icmlauthorlist}

\icmlauthor{Agnideep Aich}{ull}
\icmlauthor{Sameera Hewage}{wl}
\icmlauthor{Md Monzur Murshed}{msu}
\icmlauthor{Bruce Wade}{ull}
\icmlauthor{Ashit Baran Aich}{ind}

\end{icmlauthorlist}

\icmlaffiliation{ull}{Department of Mathematics, University of Louisiana at Lafayette, Lafayette, Louisiana, USA}
\icmlaffiliation{wl}{Department of Physical Sciences \& Mathematics, West Liberty University, West Liberty, West Virginia, USA}
\icmlaffiliation{msu}{Department of Mathematics and Statistics, Minnesota State University, Mankato, Minnesota, USA}
\icmlaffiliation{ind}{Department of Statistics, formerly of Presidency College, Kolkata, India}

\icmlcorrespondingauthor{Agnideep Aich}{agnideep.aich1@louisiana.edu}
\icmlcorrespondingauthor{Ashit Baran Aich}{aichnsou@gmail.com}

  % You may provide any keywords that you find helpful for describing your
  % paper; these are used to populate the "keywords" metadata in the PDF but
  % will not be shown in the document
  \icmlkeywords{Machine Learning, ICML}

  \vskip 0.3in
]

% this must go after the closing bracket ] following \twocolumn[ ...

% This command actually creates the footnote in the first column listing the
% affiliations and the copyright notice. The command takes one argument, which
% is text to display at the start of the footnote. The \icmlEqualContribution
% command is standard text for equal contribution. Remove it (just {}) if you
% do not need this facility.

% Use ONE of the following lines. DO NOT remove the command.
% If you have no special notice, KEEP empty braces:
\printAffiliationsAndNotice{}  % no special notice (required even if empty)
% Or, if applicable, use the standard equal contribution text:
% \printAffiliationsAndNotice{\icmlEqualContribution}==============================================================================
% ABSTRACT
% ==============================================================================

\begin{abstract}
Modern neural networks often produce miscalibrated confidence scores and struggle to detect out-of-distribution (OOD) inputs, while most existing methods post-process outputs without testing internal consistency. We introduce the Bag-of-Coins (BoC) probe, a non-parametric diagnostic of logit coherence that compares softmax confidence $\hat p$ to an aggregate of pairwise Luce-style dominance probabilities $\bar q$, yielding a deterministic coherence score and a p-value-based structural score. Across ViT, ResNet, and RoBERTa with ID/OOD test sets, the coherence gap $\Delta=\bar q-\hat p$ reveals clear ID/OOD separation for ViT (ID ${\sim}0.1$-$0.2$, OOD ${\sim}0.5$-$0.6$) but substantial overlap for ResNet and RoBERTa (both ${\sim}0$), indicating architecture-dependent uncertainty geometry. As a practical method, BoC improves calibration only when the base model is poorly calibrated (ViT: ECE $0.024$ vs.\ $0.180$) and underperforms standard calibrators (ECE ${\sim}0.005$), while for OOD detection it fails across architectures (AUROC $0.020$-$0.253$) compared to standard scores ($0.75$-$0.99$). We position BoC as a research diagnostic for interrogating how architectures encode uncertainty in logit geometry rather than a production calibration or OOD detection method.
\end{abstract}

% ==============================================================================
% 1. INTRODUCTION
% ==============================================================================
% Updated Introduction and Related Work sections with additional citations

\section{Introduction}
\label{sec:introduction}

The successful deployment of deep neural networks in critical domains, from medical diagnostics to autonomous navigation, hinges not only on their predictive accuracy but also on the reliability of their confidence estimates. A model that is ``correctly uncertain'' is often more valuable than one that is ``confidently wrong.'' However, it is well established that modern classifiers, particularly those with high capacity, are often poorly calibrated, systematically over- or under-estimating the true likelihood of their predictions being correct \citep{guo2017calibration}.  This deficiency poses a significant risk, motivating a broad search for effective calibration techniques and methods to detect when models encounter out-of-distribution (OOD) inputs.

Recent work has moved beyond simple histogram-based metrics by proposing differentiable calibration objectives and meta-learning frameworks that can be optimized during training.  For example, \citet{wang2024meta} introduce a meta-regularization framework that couples a sample-wise \emph{gamma} network with a smooth expected calibration error surrogate to learn better calibrated models; their method reduces bias in ECE computations by replacing binning with a Gaussian-kernel estimator and jointly learns focal-loss weights to regularize the backbone network.  Similarly, \citet{bohdal2023meta} propose a differentiable expected calibration error and a meta-learning strategy that directly optimizes validation-set calibration quality.  These innovations highlight ongoing efforts to integrate calibration objectives into the learning process rather than treat calibration as a purely post-hoc procedure.

Current confidence estimation and OOD detection approaches largely fall into two categories.  \textit{Post-hoc calibration} methods, such as temperature scaling \citep{guo2017calibration}, fit a simple parametric transformation of the output logits on a held-out validation set.  More flexible non-parametric methods like isotonic regression \citep{zadrozny2002transforming} offer higher expressive power at the cost of additional data requirements.  For OOD detection, methods such as maximum softmax probability, energy-based scores \citep{liu2020energy}, ODIN \citep{liang2018odin} and Mahalanobis distance \citep{lee2018simple} have shown strong empirical performance.  A second category involves modifying the training process or architecture itself: Bayesian neural networks and deep ensembles \citep{lakshminarayanan2017simple}, Monte Carlo dropout \citep{gal2016dropout}, and more recent meta-regularization frameworks \citep{wang2024meta} explicitly optimize calibration during training.  Outside of calibration, training-driven OOD detectors have been proposed to enlarge the margin between ID and OOD scores: the margin-bounded confidence score (MaCS) \citep{tamang2025macs} introduces a constraint that penalizes high confidence on OOD inputs and significantly improves detection performance, while extended logit normalization \citep{elognorm2025} incorporates feature distance-awareness into LogitNorm to provide robust separability and hyperparameter-free tuning.

A fundamental limitation of most existing methods is that they treat the logit vector as a fixed object and seek to transform or threshold it, rather than interrogating its internal structure for evidence of reliability.  Recent work has begun to explore the potential of the logit space itself.  In particular, \citet{liang2025logitgap} proposed \emph{LogitGap}, a post-hoc OOD detector that exploits the relationship between the maximum logit and the remaining logits to enhance separation between ID and OOD samples. Other authors have observed that pre-trained vision transformers can separate distributions in logit space, motivating further analysis of logit geometry.  Motivated by random utility theory, we ask a different question: does the structure of the logit vector itself reveal how different architectures encode uncertainty?  To answer this, we introduce the Bag-of-Coins (BoC) test, a simple, non-parametric statistical probe applied directly to the logits of a single prediction.  Our method requires no additional data or training.  Grounded in the principles of random utility theory, which provides a theoretical link between softmax probabilities and pairwise comparisons, we examine whether a model's stated softmax confidence is coherent with its internal pairwise logit geometry by repeatedly pitting the top-ranked class against random competitors.

Our empirical investigation across three modern architectures, namely, Vision Transformers (ViT), ResNet, and RoBERTa, reveals marked differences in how logit coherence behaves under distribution shift. ViT exhibits clear separation between in-distribution (ID) and out-of-distribution (OOD) samples in the coherence gap $\Delta=\bar q-\hat p$, with ID samples concentrated around $\Delta\sim 0.1$--$0.2$ while OOD samples shift to $\Delta\sim 0.5$--$0.6$. In contrast, ResNet and RoBERTa exhibit substantially weaker separation in $\Delta$ in our benchmarks: both show a dominant mass near $\Delta\approx 0$ with heavy overlap between ID and OOD, with OOD distributions appearing broader (especially for ResNet) but not yielding a clean coherence-based discriminator as in ViT.

However, this structural difference does not translate into practical utility.  As a calibration method, BoC improves over uncalibrated softmax only when the base model is poorly calibrated (ViT on CIFAR-10: ECE $0.024$ vs. $0.180$), but underperforms standard post-hoc methods (which achieve ECE $\sim 0.005$) and degrades performance when applied to already well-calibrated models (ResNet, RoBERTa).  For OOD detection, BoC fails catastrophically across all architectures under our binomial-tail scoring convention (AUROC $0.020$–$0.253$, far below random chance), while established methods (Energy, MSP, Mahalanobis) achieve AUROC $0.75$–$0.99$.  These failures are particularly striking for ViT, where clear coherence-based distributional signals exist but are not effectively leveraged by our p-value-based scoring.

We therefore position BoC not as a production-ready calibration or OOD detection method, but as a research diagnostic tool that reveals how different architectures encode uncertainty in their logit geometry.  We found a clear difference: ViT shows strong separation between ID and OOD in coherence structure in our tests, while ResNet and RoBERTa show much weaker separation. This suggests that to understand model reliability, we need to look at the internal geometric structure of predictions, not just surface statistics.
Our contribution is a principled, statistically grounded probe that enables such examination and opens directions for architecture-aware uncertainty quantification.

\textbf{Paper organization.} Section~\ref{sec:related_work} reviews related work; Section~~\ref{sec:notation} introduces notation; Section~\ref{sec:quality_measures} defines our confidence-quality measures; Section~\ref{sec:boc_method} presents the Bag-of-Coins (BoC) probe; Section~\ref{sec:theoretical_foundation} establishes its theoretical foundation and statistical validity; Section~\ref{sec:experiments} reports experiments; and, Section~\ref{sec:conclusion} concludes with limitations and future directions.

\section{Related Work}
\label{sec:related_work}

Our work intersects with several established lines of research in model reliability, calibration, and uncertainty quantification.

\textbf{Post-hoc calibration.}  A large body of work has studied how to correct the calibration of a trained classifier without modifying its parameters.  Platt scaling and its multi-class generalization, temperature scaling, fit one or two parameters to rescale logits before the softmax operation \citep{platt1999probabilistic,guo2017calibration}.  Isotonic regression learns a flexible monotonic transformation on held-out data \citep{zadrozny2002transforming}.  Vector scaling and Dirichlet calibration apply class-specific affine transformations to logits and have been shown to reduce calibration error on large benchmarks.  More recently, researchers have sought to incorporate calibration objectives into the training process itself.  \citet{wang2024meta} propose a meta-regularization framework that learns sample-wise weighting for Focal loss and optimizes a smooth expected calibration error surrogate, achieving unbiased calibration improvements across multiple datasets.  \citet{bohdal2023meta} introduce a differentiable expected calibration error and a meta-learning procedure that directly optimizes calibration on a validation set. These methods blur the line between post-hoc and training-time calibration, highlighting the growing interest in differentiable calibration objectives.

\textbf{Architectural and training-based methods.}  Some approaches aim to build models that are inherently better calibrated or better able to distinguish ID from OOD inputs.  Deep ensembles \citep{lakshminarayanan2017simple} average predictions across multiple independently trained networks to capture epistemic uncertainty, while Monte Carlo dropout \citep{gal2016dropout} interprets dropout as approximate Bayesian inference.  The recent meta-regularization framework of \citet{wang2024meta} combines a smooth calibration surrogate with sample-wise weights, further improving calibration.  For OOD detection, several methods modify the training objective to enlarge the gap between ID and OOD scores.  Margin-bounded confidence scores (MaCS) \citep{tamang2025macs} penalize high confidence on OOD data and enforce a margin between ID and OOD confidence during fine tuning.  Extended Logit Normalization (ELogitNorm) \citep{elognorm2025} incorporates feature distance-awareness into the LogitNorm loss to improve OOD separability and calibration without introducing hyperparameters.  These training-driven approaches complement traditional post-hoc scores by altering the representation or loss so that OOD samples are more separable at inference time.

\textbf{Logit analysis for OOD detection.}  A growing line of work recognizes that logits contain richer information than softmax probabilities, particularly for detecting distributional shift.  Energy-based methods use the log-sum-exp of logits as an OOD score \citep{liu2020energy}, while Mahalanobis distance computes the distance between test logits and class-conditional logit centroids \citep{lee2018simple}.  Recent papers explicitly analyze the configuration of the logit vector.  \citet{liang2025logitgap} propose \emph{LogitGap}, a post-hoc detector that examines the gap between the maximum logit and the remaining logits and shows that this gap provides strong separation between ID and OOD samples across vision-language and vision-only models.  Margin-bounded confidence scores and extended logit normalization can also be viewed as manipulating the logit space to improve separability.  Our Bag-of-Coins probe is distinct in that it leverages a frequentist hypothesis test to assess the \emph{internal coherence} of the logit vector rather than its magnitude or location. Inspired by random utility theory, we test whether the top logit's dominance over randomly chosen competitors is consistent with the model's softmax confidence, yielding a non-parametric diagnostic of Luce/softmax coherence.

\textbf{Surveys and broader context.}  Comprehensive reviews have recently examined the rapidly evolving landscape of OOD detection and calibration.  \citet{lu2024survey} present a task-oriented survey of OOD detection methods and classify them as training-driven or training-agnostic, noting the growing importance of pre-trained models and non-traditional scenarios such as test-time adaptation and multi-modal data. These surveys underscore the need for diagnostic tools that can illuminate architectural differences and inform the development of future methods.  Our work contributes to this broader conversation by providing a simple probe that uncovers how different model architectures encode uncertainty in their logit geometry.

% ==============================================================================
% NOTATION
% ==============================================================================
\section{Notation}
\label{sec:notation}

In this section, we introduce the notation used throughout the paper. We consider a multi-class classification problem with the following symbols: $\mathcal{X}, \mathcal{Y}, C, d, f, x, z, \hat{y}, \hat{p}, \sigma(\cdot), k, W, p_{\text{val}}, c_{\text{BoC}}, s_{\text{BoC}}, \Delta,$ and $p_{\text{dom}}$.

Let $\mathcal{X} \in \mathbb{R}^d$ be the input space and $\mathcal{Y} = \{1, 2, \dots, C\}$ be the set of classes. A classifier $f$ maps an input $x \in \mathcal{X}$ to a vector of logits $z \in \mathbb{R}^C$. The predicted class is $\hat{y} = \arg\max_c z_c$, and the associated maximum softmax probability (confidence) is $\hat{p} = \max_c \sigma(z)_c$, where $\sigma(z)_c=\exp(z_c)/\sum_i \exp(z_i)$.

For our Bag-of-Coins quantities, define the pairwise Luce probability between the top class and a competitor $j\neq \hat y$ as $\pi_{\hat y \succ j} = e^{z_{\hat y}}/(e^{z_{\hat y}} + e^{z_j})$. The \emph{deterministic BoC score} is the average pairwise win probability of the top class, $\bar q = \frac{1}{C-1}\sum_{j\neq \hat y} \pi_{\hat y \succ j}$, and we use it as our calibration score $c_{\text{BoC}} := \bar q$.

For hypothesis-testing and OOD scoring, we consider a randomized variant that produces an observed win count $W$ from $k$ trials by sampling competitors uniformly and drawing Bernoulli wins with success probability $\pi_{\hat y \succ j}$ (Algorithm~\ref{alg:MCV}). The associated (randomized) p-value is $p_{\text{val}} = \Pr\{\mathrm{Binomial}(k,\hat p)\ge W\}$. For stable scoring and visualization, we also use a deterministic low-variance proxy based on the expected win count $W^\star=\mathrm{round}(k\,\bar q)$: $p_{\text{val}}^\star = \Pr\{\mathrm{Binomial}(k,\hat p)\ge W^\star\}$. For OOD evaluation we use the score $s_{\text{BoC}} := 1 - p_{\text{val}}^\star$ so that higher values indicate in-distribution. We track the \emph{coherence gap} $\Delta := \bar q - \hat p$. For compatibility with prior text, we identify $p_{\text{dom}}$ with the average dominance probability and set $p_{\text{dom}} := \bar q$.

% ==============================================================================
% 2. MEASURING CLASSIFIER CONFIDENCE
% ==============================================================================
\section{Measuring Classifier Confidence}
\label{sec:quality_measures}

In this section, we formally define the quantities of interest for evaluating the reliability of a probabilistic classifier. We consider a standard multi-class classification setting. Let $\mathcal{X} \in \mathbb{R}^d$ be the input space and $\mathcal{Y} = \{1, 2, \dots, C\}$ be the set of $C$ classes. A classifier $f: \mathcal{X} \to \mathbb{R}^C$ maps an input $x \in \mathcal{X}$ to a vector of real-valued logits, $z(x) = (z_1, \dots, z_C)$.

The predicted class, $\hat{y}$, is the class with the highest logit value: $\hat{y} = \arg\max_c z_c(x)$. The model's confidence in this prediction, $\hat{p}$, is typically derived by applying the softmax function, $\sigma(\cdot)$, to the logit vector:
\begin{align}
    \hat{p} = \max_c \sigma(z(x))_c, \quad \text{where} \quad \sigma(z)_c = \frac{\exp(z_c)}{\sum_{i=1}^C \exp(z_i)}.
\end{align}
Our goal is to assess how well this confidence score $\hat{p}$, or any other derived confidence score, reflects the true correctness probability. We evaluate this along two primary axes: calibration and out-of-distribution detection.

\subsection{Confidence Calibration}
Perfect calibration requires that, for any confidence value $p \in [0, 1]$, a prediction with confidence $\hat{p} = p$ is correct with probability $p$. Formally:
\begin{align}
    \Pr(\hat{Y} = Y \mid \hat{P} = p) = p, \quad \forall p \in [0, 1].
\end{align}
We quantify deviations from this ideal using the Expected Calibration Error (ECE).

\begin{definition}[Expected Calibration Error (ECE)]
The ECE is the expectation of the difference between a model's average confidence and its accuracy within binned confidence intervals. The interval $[0,1]$ is partitioned into $M$ bins $B_m$. The ECE is defined as:
\begin{align*}
\mathrm{ECE}=\sum_{m=1}^{M}\frac{|B_m|}{N}\left|\mathrm{acc}(B_m)-\mathrm{conf}(B_m)\right|.
\end{align*}
where $N$ is the total number of samples, $|B_m|$ is the number of samples in bin $m$, and $\mathrm{acc}(B_m)$ and $\mathrm{conf}(B_m)$ are the average accuracy and average confidence of the samples in bin $B_m$, respectively. A lower ECE indicates better calibration.
\end{definition}

Unless stated otherwise, we compute ECE using $c_{\text{BoC}}=\bar q$ for BoC (and $\hat p$ for MSP), with $M=15$ equal-width bins.

\subsection{Out-of-Distribution Detection}
A reliable model should not only be well-calibrated on in-distribution (ID) data but should also exhibit low confidence when presented with out-of-distribution (OOD) data. This is a binary classification task: can the model's confidence score distinguish between ID and OOD samples?

We evaluate OOD detection using the Area Under the Receiver Operating Characteristic Curve (AUROC). For BoC we use the score $s_{\text{BoC}} := 1 - p_{\text{val}}^\star$ so that higher values indicate in-distribution (matching the convention for MSP/Energy). AUROC is computed directly on $s_{\text{BoC}}$.

Having established these formal measures of quality, we are now equipped to introduce our proposed method for generating a confidence score designed to probe the internal structure of the model's predictions.

% ==============================================================================
% 5. THE BAG-OF-COINS PROBE
% ==============================================================================
\section{The Bag-of-Coins Probe}
\label{sec:boc_method}

In this section, we introduce our method, the Bag-of-Coins (BoC) test. Unlike methods that transform model outputs, BoC is a non-parametric statistical probe that examines the \textit{internal consistency} of a single prediction by interrogating its logit vector, $z(x)$.

The core idea is to postulate a condition for ideal confidence representation. The softmax probability, $\hat{p} = \max_c \sigma(z(x))_c$, is the model's primary claim about the likelihood that its prediction $\hat{y}$ is correct. We propose that in a well-structured and internally consistent model, this external claim should be reflected in the internal geometry of its logits. Specifically, the top logit $z_{\hat{y}}$ should dominate randomly chosen competitor logits at a rate equal to $\hat{p}$. The BoC test is designed to measure a model's adherence to this property.

\textbf{Variant (Monte--Carlo test; optional).}
For completeness, we also considered a stochastic test that yields the same expectation but introduces sampling noise. It is not used for our main calibration curves.

\begin{definition}[The BoC Probe for Logit Coherence]
Given logits $z\in\mathbb{R}^C$, predicted class $\hat y$ and confidence $\hat p=\max_c\sigma(z)_c$, define the average pairwise dominance probability
\begin{align*}
\bar q=\frac{1}{C-1}\sum_{j\neq \hat y}\frac{e^{z_{\hat y}}}{e^{z_{\hat y}}+e^{z_j}}.
\end{align*}
We use $c_{\mathrm{BoC}}=\bar q$ as a deterministic confidence score for calibration.

To probe internal coherence, we consider the desideratum (null) $H_0:\bar q=\hat p$. In the Monte--Carlo variant, we obtain an observed win count $W$ from $k$ trials by sampling a competitor $j$ uniformly from $\{1,\dots,C\}\setminus\{\hat y\}$ and drawing $B\sim\mathrm{Bernoulli}(\pi_{\hat y\succ j})$ (Algorithm~\ref{alg:MCV}), then set
\begin{align*}
p_{\mathrm{val}}=\Pr\{\mathrm{Binomial}(k,\hat p)\ge W\}.
\end{align*}
For stable scoring and visualization, we also use the deterministic proxy $W^\star=\mathrm{round}(k\,\bar q)$ and define
\begin{align*}
& p_{\mathrm{val}}^\star=\Pr\{\mathrm{Binomial}(k,\hat p)\ge W^\star\},\\
& s_{\mathrm{BoC}}:=1-p_{\mathrm{val}}^\star.
\end{align*}
\end{definition}

A low $p_{\text{val}}$ indicates a rejection of the null hypothesis, suggesting the model’s logit structure is inconsistent with its softmax output (a state we term “confident delusion”).
Small $p_{\text{val}}$ indicates that the observed dominance is unusually large relative to $\hat p$ under $H_0$ (“confident delusion”); the gap $\Delta=\bar q-\hat p$ provides a signed coherence diagnostic. For calibration we use $c_{\text{BoC}}=\bar q$. For OOD scoring and logit-coherence diagnostics we use $s_{\text{BoC}}=1-p^\star_{\text{val}}$ (and $p_{\text{val}}$ when using the randomized variant).

% ==============================================================================
% 6. THEORETICAL FOUNDATION AND STATISTICAL VALIDITY
% ==============================================================================
\section{Theoretical Foundation and Statistical Validity}
\label{sec:theoretical_foundation}

In this section, we present the theoretical motivation for our test, connecting it to the principles of random utility theory. We then confirm the statistical validity of our procedure.

\subsection{Motivation from Random Utility Theory}
Random utility theory provides a standard interpretation of softmax and pairwise comparisons. Under i.i.d.\ Gumbel perturbations of logits, the softmax probability $p_c=\sigma(z)_c$ equals the probability that class $c$ attains the maximum perturbed utility among all $C$ classes. Moreover, for any pair of classes $(c,j)$, the probability that $c$ beats $j$ is the Luce (pairwise softmax) probability
\begin{align*}
\pi_{c\succ j}=\frac{e^{z_c}}{e^{z_c}+e^{z_j}}.
\end{align*}
BoC aggregates these pairwise margins for the predicted class $\hat y$ via
\begin{align*}
\bar q=\frac{1}{C-1}\sum_{j\neq \hat y}\pi_{\hat y \succ j},
\end{align*}
and uses the coherence gap $\Delta=\bar q-\hat p$ (together with a binomial-tail score) to probe whether the model's stated softmax confidence $\hat p$ is reflected in the internal pairwise geometry of its logits.

\subsection{Statistical Validity}
We analyze the Monte--Carlo variant (Algorithm~\ref{alg:MCV}), which produces an observed win count $W=\sum_{i=1}^k B_i$ from $k$ independent trials. Conditional on the sampled competitors, the trials are Bernoulli with potentially different success probabilities, so $W$ is generally Poisson--binomial rather than exactly binomial. 

We define the internal-coherence null as $H_0:\bar q=\hat p$ and compute the one-sided binomial-tail p-value
\begin{align*}
p_{\mathrm{val}}=\Pr\{\mathrm{Binomial}(k,\hat p)\ge W\}.
\end{align*}
A classical result of Hoeffding (1956) implies that, among all Poisson--binomial sums with a fixed mean, the binomial distribution maximizes the upper tail. As a consequence, the above binomial-tail p-value is super-uniform under $H_0$, yielding a finite-sample validity guarantee (Proposition~\ref{prop:valid}).

\subsection{Assumptions, Validity, and Concentration}
\label{subsec:validity}

We formalize the setting and provide finite-sample validity guarantees for the BoC $p$-value.

\textbf{Assumptions.}
(A1) \emph{Uniform competitor sampling.} In the Monte--Carlo variant, each trial independently samples $J_i$ uniformly from $\mathcal{C}=\{1,\dots,C\}\setminus\{\hat y\}$.\\
(A2) \emph{Fixed logits.} Conditioning on the input $x$, the logit vector $z(x)$ (hence $\pi_{\hat y \succ j}$ and $\hat p$) is treated as fixed across the $k$ trials.\\
(A3) \emph{Null (Luce/softmax coherence).} For $\pi_{\hat y \succ j}=\frac{e^{z_{\hat y}}}{e^{z_{\hat y}}+e^{z_j}}$ and $\bar q=\frac{1}{C-1}\sum_{j\neq \hat y}\pi_{\hat y \succ j}$, the null is $H_0:\bar q=\hat p$, where $\hat p=\max_c \sigma(z)_c$.

Let $B_i\mid (J_i{=}j)\sim\mathrm{Bernoulli}(\pi_{\hat y \succ j})$ be the win indicator at trial $i$ and $W=\sum_{i=1}^k B_i$.

\begin{lemma}[Binomial tail upper-bounds Poisson--binomial tail]
\label{lem:binom_upper}
Let $X=\sum_{i=1}^k Y_i$ where $\{Y_i\}$ are independent $\mathrm{Bernoulli}(p_i)$ with mean $\mu=\sum_{i=1}^k p_i$. Let $Z\sim \mathrm{Binomial}(k,\mu/k)$. Then for all integers $t$,
\begin{align*}
\Pr\{X\ge t\}\;\le\;\Pr\{Z\ge t\}.
\end{align*}
\end{lemma}

Proof is given in Appendix~\ref{app:proofs}.

\begin{proposition}[Finite-sample $p$-value validity]
\label{prop:valid}
Under \textup{(A1)--(A3)}, let $W=\sum_{i=1}^k B_i$ and define
\begin{align*}
p_{\mathrm{val}}\;=\;\Pr\!\big\{\mathrm{Binomial}(k,\hat p)\ge W\big\}.
\end{align*}
Then for all $\alpha\in(0,1)$,
\begin{align*}
\Pr\big(p_{\mathrm{val}}\le \alpha\big)\;\le\;\alpha.
\end{align*}
That is, $p_{\mathrm{val}}$ is (finite-sample) valid.
\end{proposition}

For full proof, see Appendix~\ref{app:proofs}.

\begin{corollary}[Concentration and choice of $k$]
\label{cor:hoeffding}
With the same setup (conditioning on $z$), for any $\varepsilon>0$,
\begin{align*}
\Pr\!\Big(\Big|\tfrac{W}{k}-\bar q\Big|\ge \varepsilon\Big)\;\le\;2\exp(-2k\varepsilon^2).
\end{align*}
In particular, $k=100$ yields $\Pr(|W/k-\bar q|\ge 0.20)\le 6.7\times 10^{-4}$, and $k=200$ yields the same bound $\le 2.1\times 10^{-7}$.
\end{corollary}

Proof is provided in Appendix~\ref{app:proofs}.

\textbf{Deterministic variant (used in plots).}
Define the deterministic BoC score $c_{\mathrm{BoC}}:=\bar q$ and the expected-wins surrogate $W^\star=\mathrm{round}(k\,\bar q)$. Set
\begin{align*}
p_{\mathrm{val}}^\star=\Pr\{\mathrm{Binomial}(k,\hat p)\ge W^\star\},
\qquad
s_{\mathrm{BoC}}:=1-p_{\mathrm{val}}^\star.
\end{align*}
Since $W$ concentrates around $W^\star$ by Corollary~\ref{cor:hoeffding} and the binomial tail is monotone in its argument, $p_{\mathrm{val}}^\star$ is a conservative, low-variance proxy for $p_{\mathrm{val}}$ that preserves the intended ranking for ID/OOD scoring while removing Monte--Carlo noise in figures. The finite-sample validity guarantee in Proposition~\ref{prop:valid} applies to the randomized p-value $p_{\mathrm{val}}$ computed from $W$; $p_{\mathrm{val}}^\star$ is used as a stable score and is not claimed to be a valid p-value.

% ==============================================================================
% 4. EXPERIMENTS
% ==============================================================================
% ==============================================================================
% EXPERIMENTS SECTION - CORRECTED AND COMPREHENSIVE
% ==============================================================================
\section{Experiments}
\label{sec:experiments}

This section empirically evaluates the Bag-of-Coins (BoC) probe across three settings: two vision architectures (ViT and ResNet) and one NLP architecture (RoBERTa). Our goal is not to claim BoC as a universally superior calibrator or OOD detector; rather, we use it as a diagnostic tool that exposes differences in how architectures represent uncertainty through their logit structures.

\textbf{Common protocol.}
Across all experiments, we report: (1) \textbf{Calibration} on in-distribution (ID) test data using Expected Calibration Error (ECE with $M{=}15$ bins), negative log-likelihood (NLL), and Brier score; and (2) \textbf{Out-of-distribution (OOD) detection} performance separating ID (positive class) from OOD (negative class) using Area Under the Receiver Operating Characteristic Curve (AUROC) and False Positive Rate at 95\% True Positive Rate (FPR@95\%TPR). Higher AUROC and lower FPR@95\%TPR indicate better OOD detection.

For BoC calibration, we use the confidence score $c_{\mathrm{BoC}}=\bar q$, where $\bar q$ is the average pairwise dominance probability defined in Section~\ref{sec:boc_method}. For OOD scoring, we use $s_{\mathrm{BoC}} = 1-p^\star_{\mathrm{val}}$, where $p^\star_{\mathrm{val}}$ is the binomial-tail p-value computed from the expected win count $W^\star=\mathrm{round}(k\,\bar q)$ (Algorithm~\ref{alg:boc}). This ensures that higher scores indicate in-distribution samples, matching the convention for MSP and Energy scores. We set $k{=}100$ as the main BoC trial count and perform ablation studies with $k\in\{20,50,100,200\}$ to compare deterministic versus Monte--Carlo BoC variants. All experiments use a fixed random seed (42) for reproducibility.

% ==============================================================================
% ==============================================================================
\subsection{Experiment 1: ViT on CIFAR-10 (ID) with SVHN (OOD)}
\label{subsec:exp_vit}

\textbf{Setup.}
We evaluate a pretrained Vision Transformer (ViT-Base/16) fine-tuned on CIFAR-10 using the model checkpoint \texttt{aaraki/vit-base-patch16-224-in21k-\\finetuned-cifar10} from HuggingFace. The ID dataset is the CIFAR-10 test set (10,000 images, 10 classes), and the OOD dataset is the SVHN test set (26,032 images). We use a validation split of 5,000 samples from the CIFAR-10 training data for fitting post-hoc calibration methods. Batch size is 128, and images are preprocessed using the ViT image processor.

\textbf{Baseline methods.}
For calibration, we compare: (i) Maximum Softmax Probability (MSP), (ii) Temperature Scaling fitted on validation data, (iii) Isotonic Regression on MSP confidence, (iv) Vector Scaling (multinomial logistic regression on logits, used as a Dirichlet-style calibration baseline), and (v) BoC using $c_{\mathrm{BoC}}=\bar q$.

For OOD detection, we compare: (i) MSP confidence, (ii) the (negative) energy score used as an ID score,
$s_{\mathrm{Energy}}=\log\sum_c\exp(z_c)$ (equivalently $-E(x)$ with $E(x)=-\log\sum_c\exp(z_c)$) using $T_E{=}1$,
(iii) ODIN with temperature $T{=}1000$ and perturbation magnitude $\epsilon{=}0.0014$,
(iv) Mahalanobis distance computed on logits using class-conditional Gaussian statistics estimated from training data,
and (v) BoC score $s_{\mathrm{BoC}}=1-p^\star_{\mathrm{val}}$ with $k{=}100$.

\textbf{Calibration results.}
Table~\ref{tab:vit_calib} reports calibration performance on the CIFAR-10 test set. The uncalibrated MSP baseline is substantially miscalibrated with ECE $0.1802$. Post-hoc calibration methods dramatically improve calibration: Temperature Scaling achieves ECE $0.0080$, while Isotonic Regression and Vector Scaling achieve ECE $0.0053$ and $0.0053$, respectively. These methods also reduce NLL (from $0.2564$ to ${\sim}0.07$) and Brier score (from $0.0481$ to ${\sim}0.016$).

BoC calibration using $c_{\mathrm{BoC}}=\bar q$ yields ECE $0.0243$, a large improvement over uncalibrated MSP but weaker than the best post-hoc calibrators. BoC achieves a competitive Brier score ($0.0189$), while its NLL remains at the uncalibrated value ($0.2564$) because BoC provides only a scalar confidence score rather than a recalibrated probability vector.

\begin{table}[t]
\centering
\caption{ViT calibration on CIFAR-10 (ID) test set. Lower is better for all metrics.}
\label{tab:vit_calib}
% Even smaller than footnotesize
\setlength{\tabcolsep}{2pt} 
\begin{tabular}{lccc}
\toprule
Method & ECE ($M{=}15$) & NLL & Brier \\
\midrule
MSP & 0.1802 & 0.2564 & 0.0481 \\
Temperature Scaling & 0.0080 & 0.0712 & 0.0160 \\
Isotonic Regression & 0.0053 & 0.2564 & 0.0155 \\
Vector Scaling (Dirichlet) & 0.0053 & 0.0691 & 0.0156 \\
BoC ($c=\bar q$) & 0.0243 & 0.2564 & 0.0189 \\
\bottomrule
\end{tabular}
\end{table}

\textbf{OOD detection.}
Table~\ref{tab:vit_ood} reports OOD detection results for CIFAR-10 (ID) versus SVHN (OOD). Energy and MSP achieve excellent performance with AUROC $0.991$ and $0.987$, respectively, and low FPR@95\%TPR ($0.050$ and $0.077$). Mahalanobis distance also performs well (AUROC $0.968$, FPR@95\%TPR $0.152$). ODIN fails in this setting with AUROC $0.497$ (near random chance) and FPR@95\%TPR $0.991$, indicating that the chosen ODIN hyperparameters are not effective for this model-dataset pair.

\textbf{BoC exhibits catastrophic failure for OOD detection}, achieving AUROC $0.020$ and FPR@95\%TPR $1.000$. This implies that under the current scoring convention $s_{\mathrm{BoC}}=1-p^\star_{\mathrm{val}}$, the BoC score assigns \emph{higher} values to OOD samples than to ID samples, inverting the desired ranking.

\begin{table}[t]
\centering
\caption{ViT OOD detection: CIFAR-10 (ID) vs.\ SVHN (OOD). Higher AUROC is better; lower FPR@95\%TPR is better.}
\label{tab:vit_ood}
\begin{tabular}{lcc}
\toprule
Method & AUROC & FPR@95\%TPR \\
\midrule
MSP & 0.9868 & 0.0766 \\
Energy & 0.9914 & 0.0498 \\
ODIN & 0.4965 & 0.9912 \\
Mahalanobis (logits) & 0.9680 & 0.1520 \\
BoC ($1-p^\star_{\mathrm{val}}$) & 0.0203 & 1.0000 \\
\bottomrule
\end{tabular}
\end{table}

\textbf{Coherence-gap diagnostics.}
Despite BoC's failure as an OOD detector under $s_{\mathrm{BoC}}$, the coherence gap $\Delta=\bar q-\hat p$ exposes a strong structural difference between ID and OOD samples: ID samples concentrate around $\Delta\approx 0.1$--$0.2$, while OOD samples shift substantially higher with most mass around $\Delta\approx 0.5$--$0.6$ (Figure~\ref{fig:vit_delta}, left; (see Appendix~\ref{app:vit_supp})). Additional diagnostic plots and trial-count ablations are deferred to Appendix~\ref{app:vit_supp}.

% ==============================================================================
% ==============================================================================
\subsection{Experiment 2: ResNet on CIFAR-10 (ID) with SVHN (OOD)}
\label{subsec:exp_resnet}

\textbf{Setup.}
We repeat the CIFAR-10 (ID) versus SVHN (OOD) evaluation using a pretrained ResNet-20 classifier
(\texttt{cifar10\_resnet20} from \texttt{chenyaofo/pytorch-cifar-models}). The protocol mirrors
Experiment~\ref{subsec:exp_vit}: CIFAR-10 test set (10,000 images) as ID, SVHN test set (26,032 images) as OOD,
a validation split of 5,000 samples from CIFAR-10 training data for fitting post-hoc calibration methods,
batch size 256, and standard CIFAR-10 normalization. We compare the same calibration methods (MSP, Temperature Scaling,
Isotonic Regression, Vector Scaling, and BoC with $c_{\mathrm{BoC}}=\bar q$) and the same OOD detection methods
(MSP, Energy, ODIN, Mahalanobis on logits, and BoC with $s_{\mathrm{BoC}}=1-p^\star_{\mathrm{val}}$).

\textbf{Calibration results.}
Table~\ref{tab:resnet_calib} shows a striking contrast with the ViT setting.
The uncalibrated MSP baseline is already reasonably calibrated with ECE $0.0390$.
In this run, all post-hoc calibrators \emph{worsen} calibration: Temperature Scaling increases ECE to $0.0559$,
Isotonic Regression to $0.0593$, and Vector Scaling to $0.0536$. These methods also increase NLL
(from $0.2815$ to $0.392$--$0.443$) and Brier score (from $0.0547$ to ${\sim}0.061$--$0.063$).

BoC calibration produces the worst ECE ($0.0695$) and the largest Brier score ($0.0702$).
Thus, for this pretrained ResNet-20 on CIFAR-10, the raw softmax confidence appears relatively well-behaved, and
post-hoc adjustments (including BoC-as-confidence) introduce additional miscalibration rather than correcting it.

\begin{table}[t]
\centering
\caption{ResNet calibration on CIFAR-10 (ID) test set. Lower is better for all metrics. Post-hoc calibration methods degrade performance in this run.}
\label{tab:resnet_calib}
\setlength{\tabcolsep}{2pt} 
\begin{tabular}{lccc}
\toprule
Method & ECE ($M{=}15$) & NLL & Brier \\
\midrule
MSP & 0.0390 & 0.2815 & 0.0547 \\
Temperature Scaling & 0.0559 & 0.4433 & 0.0619 \\
Isotonic Regression & 0.0593 & 0.2815 & 0.0630 \\
Vector Scaling (Dirichlet) & 0.0536 & 0.3925 & 0.0612 \\
BoC ($c=\bar q$) & 0.0695 & 0.2815 & 0.0702 \\
\bottomrule
\end{tabular}
\end{table}

\textbf{OOD detection.}
Table~\ref{tab:resnet_ood} reports OOD detection for CIFAR-10 (ID) versus SVHN (OOD).
Energy performs best with AUROC $0.906$ and FPR@95\%TPR $0.547$, followed closely by ODIN (AUROC $0.893$, FPR@95\%TPR $0.572$).
MSP achieves moderate performance (AUROC $0.875$, FPR@95\%TPR $0.705$), while Mahalanobis is weakest among baselines
(AUROC $0.821$, FPR@95\%TPR $0.805$).

\textbf{BoC again fails for OOD detection}, achieving AUROC $0.126$ (well below random chance) and FPR@95\%TPR ${\approx}1.0$.
This indicates that under the scoring convention $s_{\mathrm{BoC}}=1-p^\star_{\mathrm{val}}$, BoC assigns systematically higher
``ID scores'' to OOD samples than to ID samples (inverted ranking).

\begin{table}[t]
\centering
\caption{ResNet OOD detection: CIFAR-10 (ID) vs.\ SVHN (OOD). Higher AUROC is better; lower FPR@95\%TPR is better.}
\label{tab:resnet_ood}
\begin{tabular}{lcc}
\toprule
Method & AUROC & FPR@95\%TPR \\
\midrule
MSP & 0.8749 & 0.7051 \\
Energy & 0.9064 & 0.5471 \\
ODIN & 0.8928 & 0.5722 \\
Mahalanobis (logits) & 0.8212 & 0.8045 \\
BoC ($1-p^\star_{\mathrm{val}}$) & 0.1262 & 0.9999 \\
\bottomrule
\end{tabular}
\end{table}

\textbf{Coherence-gap diagnostics.}
Figure~\ref{fig:resnet_delta} (see Appendix~\ref{app:resnet_supp}) highlights a key difference from ViT.
ID samples have $\Delta=\bar q-\hat p$ sharply concentrated near $0$, while OOD samples exhibit a broader distribution with a heavy tail
(up to ${\sim}0.5$), but overlap near $\Delta\approx 0$ remains substantial. Additional diagnostic plots and trial-count ablations are deferred to Appendix~\ref{app:resnet_supp}.

% ==============================================================================
% ==============================================================================
% EXPERIMENT 3: RoBERTa on AG News (ID) with DBPedia (OOD)
% ==============================================================================
% ==============================================================================
\subsection{Experiment 3: RoBERTa on AG News (ID) with DBPedia (OOD)}
\label{subsec:exp_nlp}

\textbf{Setup.}
We evaluate a 4-class RoBERTa-base classifier fine-tuned on AG News using the checkpoint
\texttt{textattack/roberta-base-ag-news}.
The in-distribution (ID) dataset is the AG News test set (7,600 articles across 4 categories),
and the out-of-distribution (OOD) dataset is the DBPedia-14 test set (70,000 Wikipedia abstracts).
We use 10\% of the AG News training data as validation (12,000 samples) for fitting post-hoc calibration methods,
and the remaining 90\% (108,000 samples) for computing training statistics (e.g., for Mahalanobis on logits).
Text is tokenized with truncation and padding; batch size is 64.

\textbf{Baseline methods.}
For calibration, we compare:
(i) Maximum Softmax Probability (MSP),
(ii) Temperature Scaling fit on validation data,
(iii) Isotonic Regression fit on validation data (applied to the top-class confidence),
(iv) Vector Scaling via multinomial logistic regression on logits (Dirichlet-style calibration),
and (v) BoC confidence using $c_{\mathrm{BoC}}=\bar q$.

For OOD detection, we compare:
(i) MSP confidence,
(ii) Energy score,
(iii) Mahalanobis score on logits,
and (iv) BoC score $s_{\mathrm{BoC}} = 1-p^\star_{\mathrm{val}}$ with $k{=}100$.
We omit ODIN for NLP because the standard ODIN perturbation relies on gradient-based input perturbations,
which are not directly meaningful in discrete token space.

\textbf{Calibration results.}
Table~\ref{tab:nlp_calib} reports calibration on the AG News test set.
The uncalibrated MSP baseline is already reasonably calibrated (ECE $=0.0260$).
In this run, all post-hoc calibrators degrade ECE: Temperature Scaling increases ECE to $0.0310$,
Isotonic Regression to $0.0330$, and Vector Scaling to $0.0305$.
BoC confidence performs worst with ECE $=0.0438$ and the largest Brier score ($0.0470$).

Importantly, Isotonic Regression and BoC here output only a \emph{scalar confidence} and do not modify the
full predictive distribution; consequently, their NLL remains equal to the uncalibrated MSP NLL
($0.1854$), while ECE/Brier reflect confidence calibration only.

\begin{table}[t]
\centering
\caption{RoBERTa calibration on AG News (ID) test set. Lower is better for all metrics.}
\label{tab:nlp_calib}
\setlength{\tabcolsep}{2pt} 
\begin{tabular}{lccc}
\toprule
Method & ECE ($M{=}15$) & NLL & Brier \\
\midrule
MSP & 0.0260 & 0.1854 & 0.0407 \\
Temperature Scaling & 0.0310 & 0.2073 & 0.0417 \\
Isotonic Regression & 0.0330 & 0.1854 & 0.0425 \\
Vector Scaling (Dirichlet) & 0.0305 & 0.2065 & 0.0415 \\
BoC ($c=\bar q$) & 0.0438 & 0.1854 & 0.0470 \\
\bottomrule
\end{tabular}
\end{table}

\textbf{OOD detection.}
Table~\ref{tab:nlp_ood} reports OOD detection for AG News (ID) versus DBPedia-14 (OOD).
This is a substantially more challenging OOD setting than CIFAR-10 vs.\ SVHN because both datasets are natural-language corpora
with potential topical overlap.
Among baselines, Energy performs best (AUROC $=0.7959$, FPR@95\%TPR $=0.6383$), followed by Mahalanobis
(AUROC $=0.7546$, FPR@95\%TPR $=0.7101$) and MSP (AUROC $=0.7481$, FPR@95\%TPR $=0.8369$).

BoC again fails as an OOD detector under the score convention $s_{\mathrm{BoC}}=1-p^\star_{\mathrm{val}}$:
it attains AUROC $=0.2531$ and FPR@95\%TPR $=0.9954$, indicating a strong inversion where many OOD samples
receive higher ``ID'' scores than ID samples.

\begin{table}[t]
\centering
\caption{RoBERTa OOD detection: AG News (ID) vs.\ DBPedia-14 (OOD). Higher AUROC is better; lower FPR@95\%TPR is better.}
\label{tab:nlp_ood}
\begin{tabular}{lcc}
\toprule
Method & AUROC & FPR@95\%TPR \\
\midrule
MSP & 0.7481 & 0.8369 \\
Energy & 0.7959 & 0.6383 \\
Mahalanobis (logits) & 0.7546 & 0.7101 \\
BoC ($1-p^\star_{\mathrm{val}}$) & 0.2531 & 0.9954 \\
\bottomrule
\end{tabular}
\end{table}

\textbf{Coherence-gap diagnostics.}
We next examine the coherence gap $\Delta=\bar q-\hat p$.
Figure~\ref{fig:nlp_delta} (see Appendix~\ref{app:nlp_supp}) shows that $\Delta$ is heavily concentrated near $0$ for both ID and OOD, with substantial overlap and only a modestly heavier OOD tail.
Additional plots (reliability/ROC) and trial-count ablations are deferred to Appendix~\ref{app:nlp_supp}.

% ==============================================================================
% SUMMARY ACROSS ALL EXPERIMENTS
% ==============================================================================
% ==============================================================================
% SUMMARY ACROSS ALL EXPERIMENTS - PARAGRAPH FORMAT
% ==============================================================================
\textbf{Summary across experiments.}
A consolidated discussion of findings across all three architectures is provided in Appendix~\ref{subsec:summary}.

% ==============================================================================
% 5. CONCLUSION
% ==============================================================================
\section{Conclusion}
\label{sec:conclusion}

We introduced the Bag-of-Coins (BoC) test as a simple, non-parametric diagnostic for probing the internal coherence of neural network predictions. BoC compares a model's declared softmax confidence $\hat p$ to a geometry-derived coherence quantity $\bar q$, summarized by the coherence gap $\Delta=\bar q-\hat p$. Across three representative architectures (ViT, ResNet, RoBERTa), this probe exposes strong architectural differences in how uncertainty is represented in logit space.

Empirically, $\Delta$ provides the clearest insight: ViT exhibits a large ID/OOD shift in coherence gaps, while ResNet and RoBERTa show substantial overlap between ID and OOD $\Delta$ distributions in our benchmarks. At the same time, the specific OOD scoring rule studied here ($s_{\mathrm{BoC}}=1-p^\star_{\mathrm{val}}$) is unreliable under our evaluation convention, producing systematically inverted rankings. For calibration, BoC can improve ECE when the base model is strongly miscalibrated, but it degrades calibration when the base model is already reasonably calibrated. These findings reinforce that BoC's current strength is interpretability and diagnosis of logit geometry, rather than a drop-in replacement for established calibrators or OOD detectors.

Promising directions for future work include: (i) designing alternative BoC-style OOD scores that exploit the coherence gap (or related geometric summaries) more directly, (ii) characterizing when and why attention-based models exhibit coherence shifts under distribution change, (iii) extending coherence diagnostics to additional modalities and modern architectures, including settings where OOD differences are semantic rather than purely distributional, and (iv) generalizing BoC from ``coins'' to ``dice'' by replacing Bernoulli trials with categorical/multinomial outcomes, yielding a \emph{Bag-of-Dice (BoD)} diagnostic for probing coherence beyond the binary/binomial setting.

\section*{Impact Statement}

Our work advances the scientific understanding of uncertainty representation in neural networks by introducing a diagnostic test (BoC) that probes internal coherence between softmax confidence and logit geometry. The primary intended use is analysis and evaluation, not deployment as a stand-alone calibration or OOD detection system.

Potential positive impacts include improved transparency in model reliability assessment and better-informed development of uncertainty quantification methods, especially in high-stakes settings where understanding confidence is important. Limitations are clearly documented: under the scoring convention evaluated in this paper, BoC does not provide competitive OOD detection performance and can degrade calibration when the base model is already well-calibrated. As with other interpretability and diagnostic tools, insights about internal structure could be misused to target models (e.g., by crafting inputs that manipulate coherence), but BoC does not directly increase model capability. We encourage using BoC alongside established evaluation practices (standard calibration metrics, robustness testing, and domain-specific safety checks) rather than as a sole reliability criterion.

%%%%%%%%%%%%%%%%%%%%%%%%%%%%%%%%%%%%%%%%%%%%%%%%%%%%%%%%%%%%%%%%%%%%%%%%%%%%%%%%%%%%

\bibliography{main}

\begin{thebibliography}{14}
\providecommand{\natexlab}[1]{#1}
\providecommand{\url}[1]{\texttt{#1}}
\expandafter\ifx\csname urlstyle\endcsname\relax
  \providecommand{\doi}[1]{doi: #1}\else
  \providecommand{\doi}{doi: \begingroup \urlstyle{rm}\Url}\fi

\bibitem[Bohdal et~al.(2023)Bohdal, Yang, and Hospedales]{bohdal2023meta}
Bohdal, O., Yang, Y., and Hospedales, T.
\newblock Meta-calibration: Learning of model calibration using differentiable expected calibration error.
\newblock \emph{Transactions on Machine Learning Research}, 2023.
\newblock Accepted by TMLR 2023.

\bibitem[Ding et~al.(2025)Ding, Liu, Unger, and Eilertsen]{elognorm2025}
Ding, Y., Liu, X., Unger, J., and Eilertsen, G.
\newblock Enhancing out-of-distribution detection with extended logit normalization.
\newblock arXiv preprint arXiv:2504.11434, 2025.

\bibitem[Gal \& Ghahramani(2016)Gal and Ghahramani]{gal2016dropout}
Gal, Y. and Ghahramani, Z.
\newblock Dropout as a bayesian approximation: Representing model uncertainty in deep learning.
\newblock In Balcan, M.~F. and Weinberger, K.~Q. (eds.), \emph{Proceedings of the 33rd International Conference on Machine Learning}, volume~48 of \emph{Proceedings of Machine Learning Research}, pp.\  1050--1059, New York, New York, USA, 2016. PMLR.

\bibitem[Guo et~al.(2017)Guo, Pleiss, Sun, and Weinberger]{guo2017calibration}
Guo, C., Pleiss, G., Sun, Y., and Weinberger, K.~Q.
\newblock On calibration of modern neural networks.
\newblock In Precup, D. and Teh, Y.~W. (eds.), \emph{Proceedings of the 34th International Conference on Machine Learning}, volume~70 of \emph{Proceedings of Machine Learning Research}, pp.\  1321--1330. PMLR, 2017.

\bibitem[Lakshminarayanan et~al.(2017)Lakshminarayanan, Pritzel, and Blundell]{lakshminarayanan2017simple}
Lakshminarayanan, B., Pritzel, A., and Blundell, C.
\newblock Simple and scalable predictive uncertainty estimation using deep ensembles.
\newblock In \emph{Advances in Neural Information Processing Systems}, pp.\  6402--6413, 2017.

\bibitem[Lee et~al.(2018)Lee, Lee, Lee, and Shin]{lee2018simple}
Lee, K., Lee, K., Lee, H., and Shin, J.
\newblock A simple unified framework for detecting out-of-distribution samples and adversarial attacks.
\newblock In \emph{Advances in Neural Information Processing Systems}, pp.\  7167--7177, 2018.

\bibitem[Liang et~al.(2025)Liang, Hou, Hu, Chang, Shan, and Chen]{liang2025logitgap}
Liang, J., Hou, R., Hu, M., Chang, H., Shan, S., and Chen, X.
\newblock Revisiting logit distributions for reliable out-of-distribution detection.
\newblock arXiv preprint arXiv:2510.20134, 2025.
\newblock Accepted to NeurIPS 2025.

\bibitem[Liang et~al.(2018)Liang, Li, and Srikant]{liang2018odin}
Liang, S., Li, Y., and Srikant, R.
\newblock Enhancing the reliability of out-of-distribution image detection in neural networks.
\newblock In \emph{International Conference on Learning Representations}, 2018.

\bibitem[Liu et~al.(2020)Liu, Wang, Jiang, and He]{liu2020energy}
Liu, W., Wang, X., Jiang, J., and He, Z.
\newblock Energy-based out-of-distribution detection.
\newblock In \emph{Advances in Neural Information Processing Systems}, pp.\  21464--21475, 2020.

\bibitem[Lu et~al.(2025)Lu, Wang, Sheng, He, Zheng, and Liang]{lu2024survey}
Lu, S., Wang, Y., Sheng, L., He, L., Zheng, A., and Liang, J.
\newblock Out-of-distribution detection: A task-oriented survey of recent advances.
\newblock \emph{ACM Computing Surveys}, 2025.
\newblock arXiv preprint arXiv:2409.11884.

\bibitem[Platt(1999)]{platt1999probabilistic}
Platt, J.
\newblock Probabilistic outputs for support vector machines and comparisons to regularized likelihood methods.
\newblock In \emph{Advances in Large Margin Classifiers}, pp.\  61--74. MIT Press, 1999.

\bibitem[Tamang et~al.(2025)Tamang, Bouadjenek, Dazeley, and Aryal]{tamang2025macs}
Tamang, L., Bouadjenek, M.~R., Dazeley, R., and Aryal, S.
\newblock Improving out-of-distribution detection by enforcing confidence margin.
\newblock \emph{Knowledge and Information Systems}, 67:\penalty0 5541--5569, 2025.
\newblock \doi{10.1007/s10115-025-02380-y}.

\bibitem[Wang \& Golebiowski(2024)Wang and Golebiowski]{wang2024meta}
Wang, C. and Golebiowski, J.
\newblock Towards unbiased calibration using meta-regularization.
\newblock \emph{Transactions on Machine Learning Research}, 2024.
\newblock arXiv preprint arXiv:2303.15057.

\bibitem[Zadrozny \& Elkan(2002)Zadrozny and Elkan]{zadrozny2002transforming}
Zadrozny, B. and Elkan, C.
\newblock Transforming classifier scores into accurate multiclass probability estimates.
\newblock \emph{ACM SIGKDD Explorations Newsletter}, 4\penalty0 (1):\penalty0 69--75, 2002.

\end{thebibliography}
\bibliographystyle{icml2026}

%%%%%%%%%%%%%%%%%%%%%%%%%%%%%%%%%%%%%%%%%%%%%%%%%%%%%%%%%%%%%%%%%%%%%%%%%%%%%%%%%%%%%%%%%%%%%%%%%
\newpage

\appendix
\onecolumn

\section{Proofs}
\label{app:proofs}

In this section, we provide the proofs of the lemma, proposition and corollary stated in the main text.

\subsection{Proof of Lemma~\ref{lem:binom_upper}}
\begin{proof}
This is a classical result due to Hoeffding (1956, Thm.\ 4): among all sums of independent Bernoulli variables with fixed mean $\mu$, the upper tail is maximized when all success probabilities are equal, i.e., $p_i\equiv \mu/k$. Equivalently, $X$ is smaller than $Z$ in convex order, which implies the stated tail inequality.\qedhere
\end{proof}

%%%%%%%%%%%%%%%%%%%%%%%%%
\subsection{Proof of Proposition~\ref{prop:valid}}

\begin{proof}
Condition on $z$ so that $\{\pi_{\hat y \succ j}\}_{j\neq \hat y}$ and $\hat p$ are fixed constants. Under (A1) and (A2), the trials are independent and
\begin{align*}
\Pr(B_i=1) &=\mathbb{E}\big[\,\Pr(B_i=1\mid J_i)\,\big] \\
&=\frac{1}{C-1}\sum_{j\neq \hat y}\pi_{\hat y \succ j}\\ &=\bar q.
\end{align*}
Hence $W$ is Poisson--binomial with parameters $\{p_i\}_{i=1}^k$ satisfying $\frac{1}{k}\sum_i p_i=\bar q$. Under $H_0$, $\bar q=\hat p$, so $W$ is Poisson--binomial with mean $k\hat p$.

Let $Z\sim\mathrm{Binomial}(k,\hat p)$. By Lemma~\ref{lem:binom_upper}, for every integer $t$,
\begin{align*}
\Pr\{W\ge t\}\;\le\;\Pr\{Z\ge t\}.
\end{align*}
Define the tail function $G(t)=\Pr\{Z\ge t\}$. Since $G$ is nonincreasing in $t$ and $p_{\mathrm{val}}=G(W)$, we have for any $\alpha\in(0,1)$,
\begin{align*}
\Pr\{p_{\mathrm{val}}\le \alpha\}
&=\Pr\{G(W)\le \alpha\}\\
&=\Pr\{W\ge G^{-1}(\alpha)\}\\
&\le \Pr\{Z\ge G^{-1}(\alpha)\}\\
&=\alpha,
\end{align*}
where $G^{-1}(\alpha):=\min\{t:\,G(t)\le \alpha\}$ and the last equality uses the definition of the quantile for a (right-)tail function. This proves super-uniformity of $p_{\mathrm{val}}$ and hence validity.\qedhere
\end{proof}

%%%%%%%%%%%%%%%%%%%%%%%%
\subsection{Proof of Corollary~\ref{cor:hoeffding}}
\begin{proof}
Given $z$, the indicators $B_i\in[0,1]$ are independent with $\mathbb{E}B_i=\bar q$. Hoeffding's inequality applies to bounded independent summands, giving
\begin{align*}
\Pr\!\left(\tfrac{1}{k}\sum_{i=1}^k B_i-\bar q\ge \varepsilon\right)\le \exp(-2k\varepsilon^2)
\end{align*}
and symmetrically for the lower tail. Union bound yields the two-sided bound.\qedhere
\end{proof}
%%%%%%%%%%%%%%%%%%%%%%%%%%%%%%%%%%%%%%%%%%%%%%%%%%%%%%%%%%%%%%%%%%%%%%%%%%%%%%%%%%%%%%
\section{Algorithms}
\label{app:algo}
In this section, we provide pseudocode for the two BoC variants referenced in the main text.

\begin{algorithm}[H]
    \caption{Bag-of-Coins (BoC): Deterministic Score and p-value}
    \label{alg:boc}
\begin{algorithmic}
    \STATE \textbf{Input:} logits $z\in\mathbb{R}^C$, trials parameter $k$ (e.g., $k{=}100$ for p-value granularity)
    \STATE Compute $p=\sigma(z)$; let $\hat y=\arg\max_c z_c$ and $\hat p=\max_c p_c$
    \STATE \textbf{Deterministic BoC score:}
           \begin{align*}
           \bar q \;=\; \frac{1}{C-1}\sum_{j\neq \hat y}\frac{e^{z_{\hat y}}}{e^{z_{\hat y}}+e^{z_j}}
           \end{align*}
    \STATE \textbf{p-value (binomial tail using expected wins):}
           \begin{align*}
           W^\star=\mathrm{round}(k\,\bar q), \qquad
           p_{\text{val}}^\star=\Pr\!\left\{\mathrm{Binomial}(k,\hat p)\ge W^\star\right\}
           \end{align*}
    \STATE \textbf{Return:} $c_{\text{BoC}}=\bar q$ \;(calibration score), \quad
           $s_{\text{BoC}}=1-p_{\text{val}}^\star$ \;(ID/OOD score)
\end{algorithmic}
\end{algorithm}

\begin{algorithm}[H]
\caption{Bag-of-Coins (Monte--Carlo variant)}
\label{alg:MCV}
\begin{algorithmic}
    \STATE \textbf{Input:} logits $z \in \mathbb{R}^C$, trials parameter $k$
    \STATE Compute $p = \sigma(z)$; let $\hat{y} = \arg\max_c z_c$ and $\hat{p} = \max_c p_c$
    \STATE $\mathcal{C} = \{1,\dots,C\} \setminus \{\hat{y}\}$
    \STATE 
    \STATE \textbf{Pairwise prob:} 
    \STATE \hspace{2em} $\displaystyle \pi_{\hat{y} \succ j} = \frac{e^{z_{\hat{y}}}}{e^{z_{\hat{y}}} + e^{z_j}}$
    \STATE 
    \STATE $W \gets 0$; \textbf{repeat} $k$ times: sample $j \sim \mathrm{Unif}(\mathcal{C})$, draw 
    \STATE \hspace{2em} $B \sim \mathrm{Bernoulli}(\pi_{\hat{y} \succ j})$, set $W \gets W + B$
    \STATE 
    \STATE $p_{\mathrm{val}} = \Pr\{\mathrm{Binomial}(k, \hat{p}) \ge W\}$
    \STATE 
    \STATE \textbf{Return:} $p_{\mathrm{val}}$
\end{algorithmic}
\end{algorithm}

\section{Supplementary Material for Experiment 1 (ViT on CIFAR-10 vs.\ SVHN)}
\label{app:vit_supp}
This section provides supplementary plots and ablations for Experiment~\ref{subsec:exp_vit}.

\begin{figure}[t]
\centering
\includegraphics[width=0.45\linewidth]{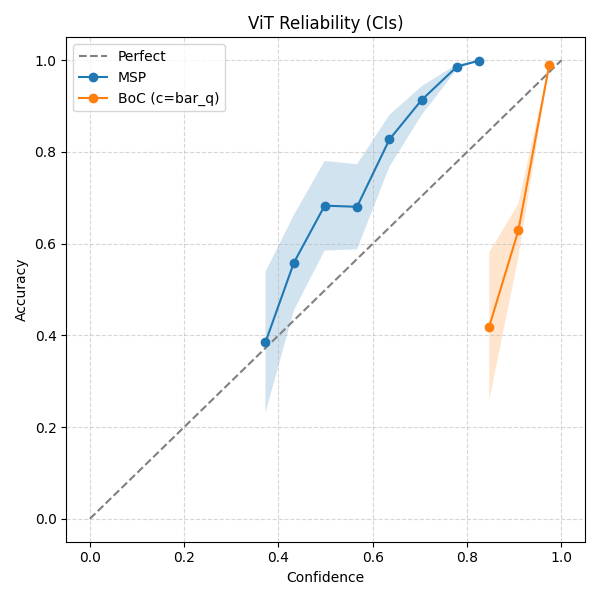}
\includegraphics[width=0.45\linewidth]{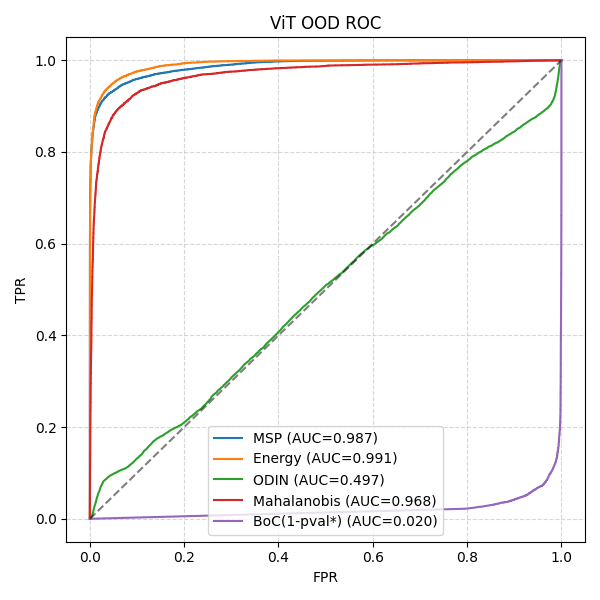}
\caption{ViT reliability diagram with bootstrap 95\% confidence intervals (left) and OOD ROC curves (right) for CIFAR-10 (ID) vs.\ SVHN (OOD).}
\label{fig:vit_rel_oodroc}
\end{figure}

\textbf{Supplementary plots.}
Figure~\ref{fig:vit_rel_oodroc} shows reliability diagrams with bootstrap confidence intervals (left) and OOD ROC curves (right) for CIFAR-10 (ID) vs.\ SVHN (OOD). MSP remains overconfident across most confidence ranges. BoC concentrates predictions into higher-confidence bins and, in those bins, remains overconfident (e.g., for confidence around $0.8$--$0.9$, the observed accuracy is markedly lower). BoC's ROC curve lies far below the diagonal, confirming inverted ranking under $s_{\mathrm{BoC}}=1-p^\star_{\mathrm{val}}$.

\begin{figure}[t]
\centering
\includegraphics[width=0.45\linewidth]{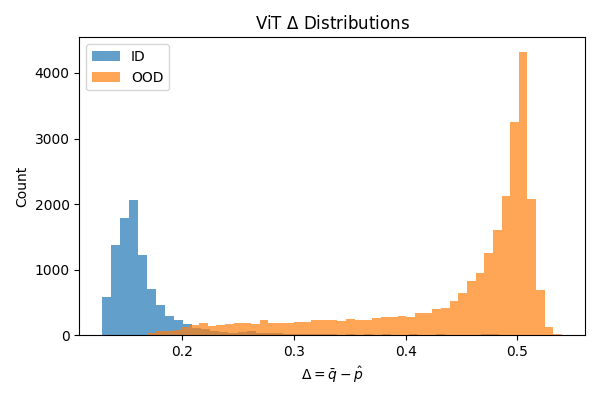}
\includegraphics[width=0.45\linewidth]{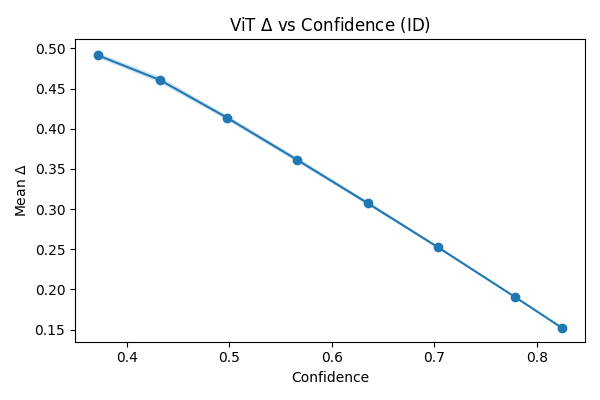}
\caption{ViT BoC coherence diagnostics: histogram of $\Delta=\bar q-\hat p$ showing clear separation between ID (blue, centered around ${\sim}0.1$--$0.2$) and OOD (orange, centered around ${\sim}0.5$--$0.6$) distributions (left), and mean $\Delta$ versus confidence on ID data showing a strong negative relationship (right).}
\label{fig:vit_delta}
\end{figure}

\textbf{Additional coherence diagnostics.}
Figure~\ref{fig:vit_delta} (right) shows mean $\Delta$ versus confidence on ID data. Mean $\Delta$ decreases steadily as confidence increases: at lower confidence (around $\hat p\approx 0.37$--$0.40$), mean $\Delta$ is approximately $0.49$, while at higher confidence (around $\hat p\approx 0.8$--$0.85$), mean $\Delta$ decreases to approximately $0.15$. This strong negative trend suggests that as the model becomes more confident on ID samples, its softmax output becomes more coherent with its internal pairwise logit structure.

\begin{figure}[t]
\centering
\includegraphics[width=0.45\linewidth]{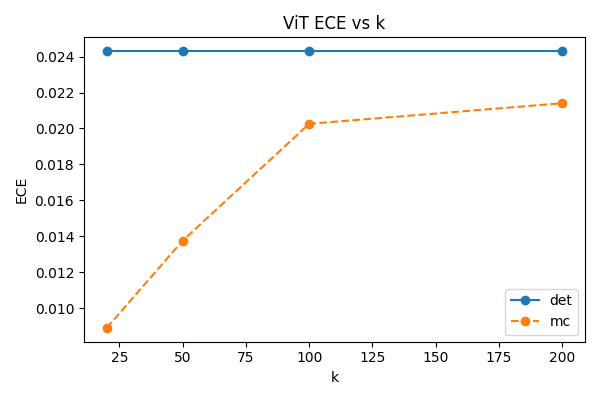}
\includegraphics[width=0.45\linewidth]{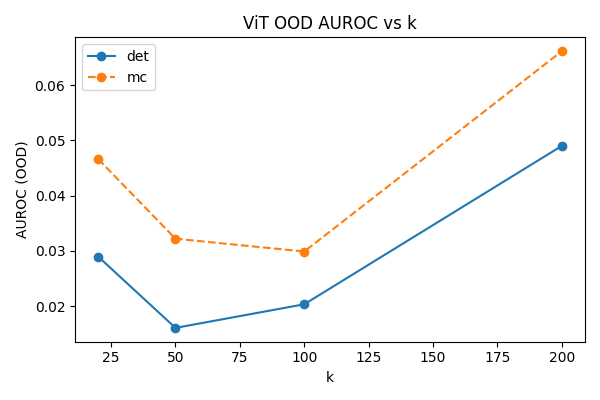}
\caption{ViT BoC sensitivity to trial count $k$: ECE versus $k$ (left) and OOD AUROC versus $k$ (right), comparing deterministic (solid) and Monte--Carlo (dashed) variants.}
\label{fig:vit_k_sweep}
\end{figure}

\textbf{Trial count ablations: deterministic versus Monte--Carlo.}
Figure~\ref{fig:vit_k_sweep} examines sensitivity to the trial count $k\in\{20,50,100,200\}$, comparing deterministic BoC (using $\bar q$) and Monte--Carlo BoC (using the sampled win-rate $\hat q$).

For calibration (left panel), deterministic BoC is stable across all $k$ (ECE stays near $0.024$). In contrast, the Monte--Carlo variant shows a clear increase in ECE as $k$ grows: it is lowest at $k{=}20$ (approximately $0.009$), then increases at $k{=}50$ (approximately $0.014$), and approaches the deterministic level by $k{=}100$--$200$ (approximately $0.020$--$0.022$). This behavior is consistent with $\hat q$ converging toward $\bar q$ as $k$ increases.

For OOD detection (right panel), both variants remain catastrophically poor (AUROC far below $0.5$ for all $k$). The AUROC values are not monotone in $k$: both curves dip around $k{=}50$--$100$ and then increase at $k{=}200$. Deterministic BoC ranges from about $0.016$ to $0.049$, while Monte--Carlo BoC ranges from about $0.030$ to $0.067$. Increasing $k$ does not remedy the fundamental inversion under $s_{\mathrm{BoC}}=1-p^\star_{\mathrm{val}}$, despite the strong $\Delta$ separation in Figure~\ref{fig:vit_delta}.

%%%%%%%%%%%%%%%%%%%%%%%%%%%%%%%%%%%%%%%%%%%%%%%%%%%%%%%%%%%%%%%%%%%%%%%%%%%%%%%%%%%%%%%

\section{Supplementary Material for Experiment 2 (ResNet on CIFAR-10 vs.\ SVHN)}
\label{app:resnet_supp}

This section provides supplementary plots and ablations for Experiment~\ref{subsec:exp_resnet}.

\begin{figure}[t]
\centering
\includegraphics[width=0.45\linewidth]{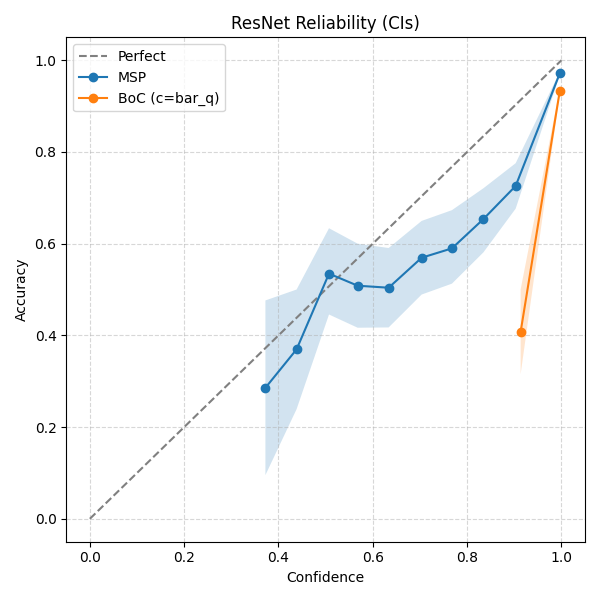}
\includegraphics[width=0.45\linewidth]{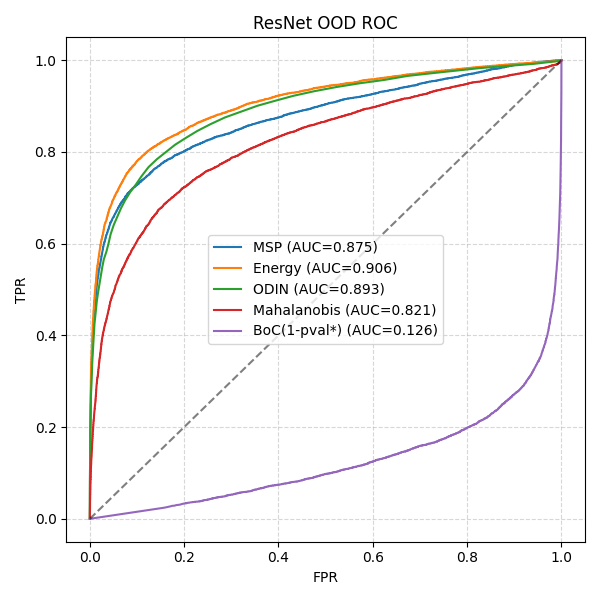}
\caption{ResNet reliability diagram with bootstrap 95\% confidence intervals (left) and OOD ROC curves (right) for CIFAR-10 (ID) vs.\ SVHN (OOD).}
\label{fig:resnet_rel_oodroc}
\end{figure}

\textbf{Supplementary plots.}
Figure~\ref{fig:resnet_rel_oodroc} shows reliability diagrams with bootstrap confidence intervals (left) and OOD ROC curves (right) for CIFAR-10 (ID) vs.\ SVHN (OOD). MSP tracks the diagonal more closely than in the ViT experiment but still shows noticeable miscalibration in mid-confidence bins. BoC places most mass in very high-confidence regions and is visibly miscalibrated there: one high-confidence bin around ${\sim}0.9$ has low observed accuracy (around ${\sim}0.4$), while the highest-confidence bin near ${\sim}1.0$ has high accuracy (around ${\sim}0.93$), indicating unstable calibration behavior across the extreme-confidence tail. BoC's ROC curve lies far below the diagonal, confirming inverted ranking under $s_{\mathrm{BoC}}=1-p^\star_{\mathrm{val}}$.

\begin{figure}[t]
\centering
\includegraphics[width=0.45\linewidth]{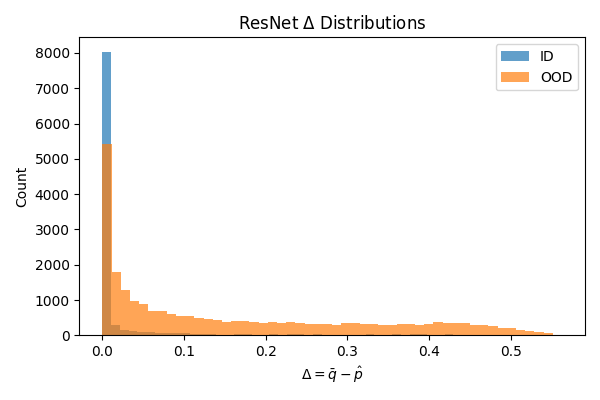}
\includegraphics[width=0.45\linewidth]{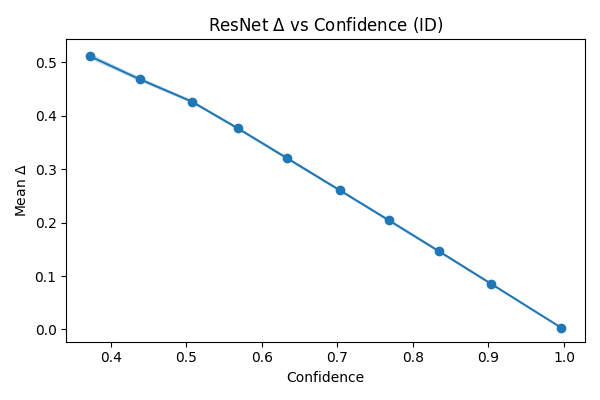}
\caption{ResNet BoC coherence diagnostics: histogram of $\Delta=\bar q-\hat p$ showing ID sharply concentrated near $0$ and OOD having a broad heavy-tailed distribution with overlap near zero (left), and mean $\Delta$ versus confidence on ID data showing a strong decreasing trend (right).}
\label{fig:resnet_delta}
\end{figure}

\textbf{Coherence-gap diagnostics.}
Figure~\ref{fig:resnet_delta} (left) highlights a key difference from ViT.
ID samples have $\Delta=\bar q-\hat p$ sharply concentrated near $0$, while OOD samples exhibit a much broader distribution with a long tail
extending to large $\Delta$ values (up to ${\sim}0.5$). However, there is still substantial overlap near $\Delta\approx 0$ because many OOD samples
also attain very small coherence gaps. Thus, unlike the ViT case where the ID/OOD $\Delta$ histograms separate cleanly, ResNet exhibits
a mixed pattern: OOD has a pronounced heavy tail in $\Delta$, but overlap near zero remains large.

Figure~\ref{fig:resnet_delta} (right) shows mean $\Delta$ versus confidence on ID data.
Mean $\Delta$ decreases approximately linearly as confidence increases: at low confidence (${ \hat p \approx 0.4}$), mean $\Delta$ is large
(around ${\sim}0.5$), and it shrinks steadily toward $0$ as $\hat p\to 1.0$. This indicates that for ResNet, high-confidence predictions tend to have
near-perfect coherence ($\bar q \approx \hat p$), while low-confidence predictions exhibit substantial coherence gaps.

\begin{figure}[t]
\centering
\includegraphics[width=0.45\linewidth]{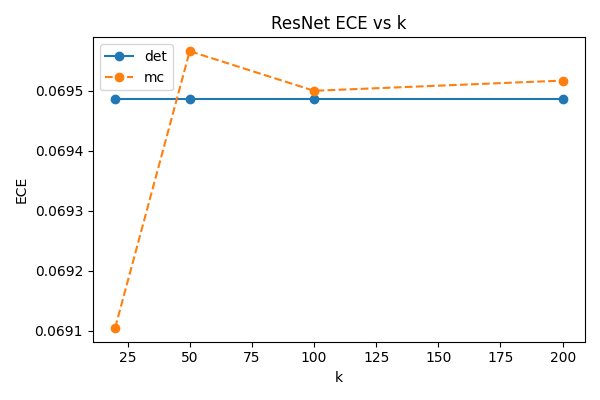}
\includegraphics[width=0.45\linewidth]{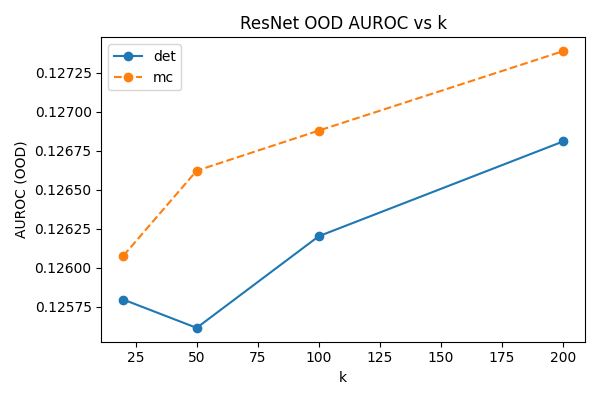}
\caption{ResNet BoC sensitivity to trial count $k$: ECE versus $k$ (left) and OOD AUROC versus $k$ (right), comparing deterministic (solid) and Monte--Carlo (dashed) variants.}
\label{fig:resnet_k_sweep}
\end{figure}

\textbf{Trial count ablations: deterministic versus Monte--Carlo.}
Figure~\ref{fig:resnet_k_sweep} shows BoC sensitivity to the trial count $k$.
For calibration (left panel), deterministic BoC is essentially constant across all $k$ (ECE ${\approx}0.0695$), as expected since it uses the analytic
$\bar q$. The Monte--Carlo variant varies only slightly (a small dip at $k{=}20$ and then values extremely close to the deterministic level), indicating
little practical sensitivity of calibration ECE to the sampling budget in this ResNet setting.

For OOD detection (right panel), both variants remain very poor for all $k$ (AUROC far below $0.5$), although AUROC increases slightly with $k$.
Deterministic BoC ranges from ${\sim}0.126$ to ${\sim}0.127$, while Monte--Carlo BoC ranges from ${\sim}0.126$ to ${\sim}0.1273$.
Thus, increasing $k$ does not resolve the inversion problem under $s_{\mathrm{BoC}}=1-p^\star_{\mathrm{val}}$.

%%%%%%%%%%%%%%%%%%%%%%%%%%%%%%%%%%%%%%%%%%%%%%%%%%%%%%%%%%%%%%%%%%%%%%%%%%%%%%%%%%%%%%%%%%%%%%%
\section{Supplementary Material for Experiment 3 (RoBERTa on AG News vs.\ DBPedia-14)}
\label{app:nlp_supp}

This section provides supplementary plots and ablations for Experiment~\ref{subsec:exp_nlp}.

\begin{figure}[t]
\centering
\includegraphics[width=0.45\linewidth]{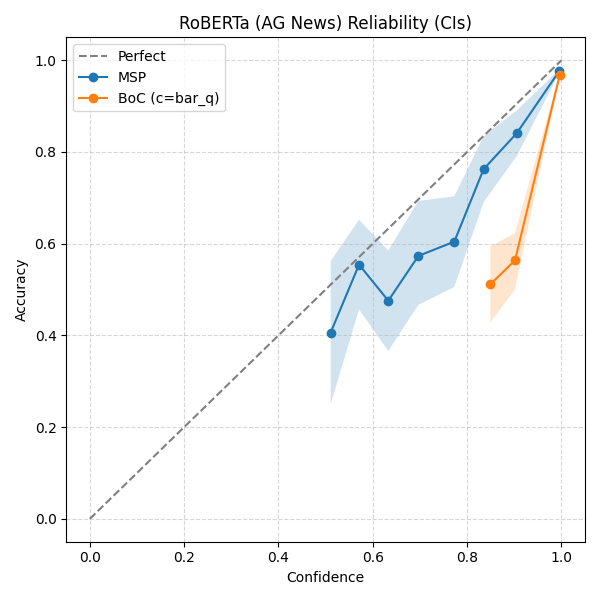}
\includegraphics[width=0.45\linewidth]{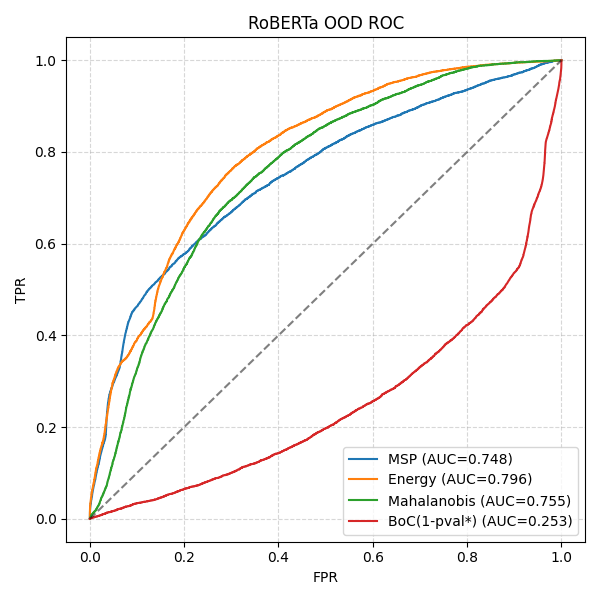}
\caption{RoBERTa reliability diagram with bootstrap 95\% confidence intervals (left) and OOD ROC curves (right) for AG News (ID) vs.\ DBPedia-14 (OOD).}
\label{fig:nlp_rel_oodroc}
\end{figure}

\textbf{Supplementary plots.}
Figure~\ref{fig:nlp_rel_oodroc} shows reliability diagrams with bootstrap confidence intervals (left) and OOD ROC curves (right) for AG News (ID) vs.\ DBPedia-14 (OOD). MSP tracks the diagonal moderately well but exhibits noise in mid-confidence bins. BoC concentrates mass in high-confidence bins and remains visibly miscalibrated, with accuracy notably below confidence around the $0.85$--$0.95$ region. The BoC ROC curve lies well below the diagonal, confirming inverted ranking under $s_{\mathrm{BoC}}=1-p^\star_{\mathrm{val}}$.

\begin{figure}[t]
\centering
\includegraphics[width=0.45\linewidth]{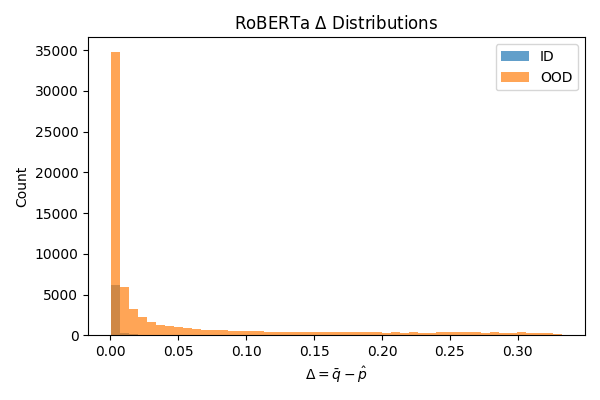}
\includegraphics[width=0.45\linewidth]{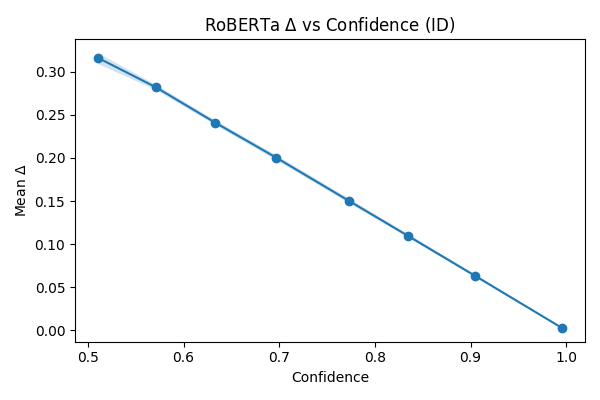}
\caption{RoBERTa BoC coherence diagnostics: histogram of $\Delta=\bar q-\hat p$ showing heavy ID/OOD overlap with both concentrated near $0$ (left), and mean $\Delta$ versus confidence on ID data (right).}
\label{fig:nlp_delta}
\end{figure}

\textbf{Coherence-gap diagnostics.}
We next examine the coherence gap $\Delta=\bar q-\hat p$.
Figure~\ref{fig:nlp_delta} (left) shows that \textbf{$\Delta$ is heavily concentrated near $0$ for both ID and OOD},
with substantial overlap and only a modestly heavier OOD tail extending to larger $\Delta$ values.
This lack of separation aligns with BoC's poor OOD performance: RoBERTa's logit geometry does not yield
a strong coherence-based signal distinguishing AG News from DBPedia in this setup.

Figure~\ref{fig:nlp_delta} (right) plots mean $\Delta$ versus confidence on ID data.
Mean $\Delta$ decreases nearly monotonically from approximately $0.31$ at confidence $\approx 0.5$
to near $0$ as confidence approaches $1.0$, indicating that high-confidence predictions are nearly coherent
($\bar q \approx \hat p$), while lower-confidence predictions exhibit larger gaps.

\begin{figure}[t]
\centering
\includegraphics[width=0.45\linewidth]{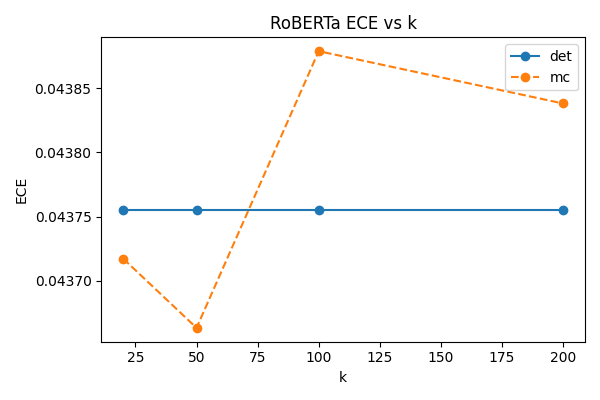}
\includegraphics[width=0.45\linewidth]{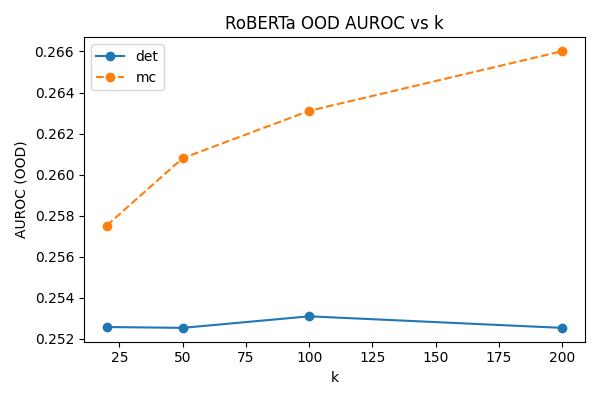}
\caption{RoBERTa BoC sensitivity to trial count $k$: ECE versus $k$ (left) and OOD AUROC versus $k$ (right), comparing deterministic (solid) and Monte--Carlo (dashed) variants.}
\label{fig:nlp_k_sweep}
\end{figure}

\textbf{Trial count ablations: deterministic versus Monte--Carlo.}
Figure~\ref{fig:nlp_k_sweep} studies sensitivity to $k\in\{20,50,100,200\}$.
For calibration (left panel), the deterministic BoC ECE is constant at $\approx 0.0438$ across all $k$,
as expected since $\bar q$ is computed analytically.
The Monte--Carlo variant shows only minor fluctuations on the order of $10^{-4}$, with the lowest ECE at $k{=}50$
and the highest around $k{=}100$; overall, changing $k$ does not materially improve calibration.

For OOD detection (right panel), deterministic AUROC remains essentially flat at $\approx 0.252$ across $k$.
Monte--Carlo AUROC increases slightly with $k$ (from $\approx 0.258$ at $k{=}20$ to $\approx 0.266$ at $k{=}200$),
but all values remain far below $0.5$, so increased trial count does not resolve the score inversion.

%%%%%%%%%%%%%%%%%%%%%%%%%%%%%%%%%%%%%%%%%%%%%%%%%%%%%%%%%%%%%%%%%%%%%%%%%%%%%%%%%%%%%
\section{Summary Across Experiments}
\label{subsec:summary}

We synthesize findings across three architectures (ViT, ResNet, RoBERTa). The main takeaway is that BoC is most useful as a \emph{diagnostic probe} of logit geometry (via the coherence gap $\Delta=\bar q-\hat p$), while the particular OOD scoring rule we used ($s_{\mathrm{BoC}}=1-p^\star_{\mathrm{val}}$) is unreliable under our evaluation convention.

\textbf{Calibration: architecture-dependent and not consistently beneficial.}
BoC confidence calibration ($c_{\mathrm{BoC}}=\bar q$) behaves differently depending on the base model's calibration. For ViT on CIFAR-10, where MSP is severely miscalibrated (ECE $0.1802$), BoC substantially improves calibration (ECE $0.0243$), though it still trails the strongest post-hoc calibrators (ECE ${\sim}0.005$--$0.008$). In contrast, for ResNet and RoBERTa---where MSP is already reasonably calibrated (ECE $0.0390$ and $0.0260$)---BoC \emph{degrades} calibration (ECE $0.0695$ and $0.0438$). Overall, standard post-hoc calibrators (temperature scaling, vector scaling, isotonic regression) are more reliable when calibration is needed, but even they can degrade performance when the base model is already well-calibrated.

\textbf{OOD detection: systematic inversion under the current BoC scoring convention.}
Under our OOD evaluation convention (where higher score indicates \emph{ID-likeness}, as for MSP confidence), the BoC OOD score $s_{\mathrm{BoC}} = 1-p^\star_{\mathrm{val}}$ is consistently \emph{inverted}: AUROC is far below chance across all three settings (ViT: $0.0203$, ResNet: $0.1262$, RoBERTa: $0.2531$), and FPR@95\%TPR is essentially $1.0$. In contrast, standard OOD baselines behave as expected, with Energy consistently strongest among them (AUROC $0.9914$ for ViT, $0.9064$ for ResNet, $0.7959$ for RoBERTa). These results indicate that, as implemented, the binomial-tail score does not provide a usable ID-likeness ranking.

\textbf{Coherence-gap diagnostics: ViT shows a strong ID/OOD shift, while ResNet and RoBERTa do not.}
Although BoC fails as an OOD detector under the current scoring, the coherence gap $\Delta=\bar q-\hat p$ reveals clear architectural differences.
ViT exhibits a pronounced distribution shift in $\Delta$: ID samples concentrate around $\Delta\approx 0.1$--$0.2$, while OOD samples shift to much larger gaps around $\Delta\approx 0.5$--$0.6$ (Figure~\ref{fig:vit_delta}). This indicates that OOD inputs disrupt ViT's internal logit-pair geometry even when softmax confidence is high.
By contrast, ResNet and RoBERTa show heavy overlap between ID and OOD $\Delta$ distributions, with both concentrated near $\Delta\approx 0$ (Figures~\ref{fig:resnet_delta}, \ref{fig:nlp_delta}). In these models, coherence-based signals appear weak for distinguishing CIFAR-10 from SVHN (ResNet) and AG News from DBPedia (RoBERTa), suggesting that logit-geometry coherence alone is insufficient for OOD separation in these settings.

\textbf{Confidence--coherence relationship: consistent within ID data across architectures.}
Across all three architectures, $\Delta$ decreases as confidence increases on ID data: low-confidence predictions exhibit larger coherence gaps, while high-confidence predictions approach $\Delta\approx 0$ (right panels of Figures~\ref{fig:vit_delta},~\ref{fig:resnet_delta},~\ref{fig:nlp_delta}). This suggests a general within-ID phenomenon: high confidence is typically accompanied by high internal coherence.

\textbf{Trial-count sensitivity: $k$ does not resolve the OOD failure mode.}
As expected, deterministic BoC is essentially insensitive to $k$ (it uses $\bar q$ analytically). Monte--Carlo BoC shows mild variation with $k$, but this does not change conclusions. In particular, for ViT the Monte--Carlo ECE \emph{increases} with $k$ (moving toward the deterministic behavior), while for ResNet and RoBERTa the Monte--Carlo calibration curve varies only slightly. For OOD detection, AUROC changes only marginally with $k$ and remains far below chance in all cases, confirming that the observed OOD failures are not due to insufficient Monte--Carlo trials.

\textbf{Takeaway.}
BoC is best viewed as a statistically grounded \emph{probe} of the relationship between softmax confidence and internal logit geometry. The coherence gap $\Delta$ reveals an architectural dichotomy: ViT shows a strong OOD-induced disruption in coherence, whereas ResNet and RoBERTa exhibit near-uniform coherence across ID and OOD in these benchmarks. Improving BoC as an OOD detector likely requires alternative scoring that leverages $\Delta$ (or related geometry) more directly than the current binomial-tail p-value rule, and may be inherently more promising in architectures where $\Delta$ actually shifts under distribution change (as in ViT).

\end{document}